\documentclass[preprint,fleqn,sort&compress,5p]{cas-dc}
\usepackage[numbers]{natbib}
\usepackage{amsmath,amsfonts}
\usepackage{array}
\usepackage{stfloats}

\begin{document}
\let\WriteBookmarks\relax
\def\floatpagepagefraction{1}
\def\textpagefraction{.001}

\shorttitle{Exemplar-based Video Colorization with Long-term Spatiotemporal Dependency}  
\shortauthors{S. Chen et al.}  

\title[mode = title]{Exemplar-based Video Colorization with Long-term Spatiotemporal Dependency}
\author[1]{Siqi Chen}
\ead{sqchen@bupt.edu.cn}
\author[2]{Xueming Li}
\ead{lixm@bupt.edu.cn}
\author[2]{Xianlin Zhang}
\ead{zxlin@bupt.edu.cn}
\author[1]{Mingdao Wang}
\ead{wmingdao@bupt.edu.cn}
\author[1]{Yu Zhang}
\ead{zhangyu\_03@bupt.edu.cn}
\author[1]{Jiatong Han}
\ead{2018213326@bupt.edu.cn}
\author[2]{Yue Zhang\corref{cor1}}
\ead{zhangyuereal@163.com}
\cortext[cor1]{Corresponding author}

\address[1]{School of Artificial Intelligence,Beijing University of Posts and Telecommunications,Beijing,102206,China}
\address[2]{School of Digital Media and Design Arts,Beijing University of Posts and Telecommunications,Beijing,102206,China}




\begin{abstract}[S U M M A R Y]
Exemplar-based video colorization is an essential technique for applications like old movie restoration. Although recent methods perform well in still scenes or scenes with regular movement, they always lack robustness in moving scenes due to their weak ability in modeling long-term dependency both spatially and temporally, leading to color fading, color discontinuity or other artifacts. To solve this problem, we propose an exemplar-based video colorization framework with long-term spatiotemporal dependency. To enhance the long-term spatial dependency, a parallelized CNN-Transformer block and a double head non-local operation are designed. The proposed CNN-Transformer block can better incorporate long-term spatial dependency with local texture and structural features, and the double head non-local operation further leverages the performance of augmented feature. While for long-term temporal dependency enhancement, we further introduce the novel linkage subnet. The linkage subnet propagate motion information across adjacent frame blocks and help to maintain temporal continuity. Experiments demonstrate that our model outperforms recent state-of-the-art methods both quantitatively and qualitatively. Also, our model can generate more colorful, realistic and stabilized results, especially for scenes where objects change greatly and irregularly.

\end{abstract}

\begin{keywords}
Video Colorization \sep Exemplar-based \sep Moving Scenes \sep Long-term Dependency \sep Spatiotemporal.
\end{keywords}

\maketitle

\section{Introduction}
Video colorization is a long-standing and highly challenging problem, it requires plausible color prediction while retaining spatiotemporal consistency, which is an essential technique for applications like old movie restoration. 

Initial video colorization methods are developed on the basis of image colorization methods. There are numerous image colorization methods \cite{Chromagan,DeepExamplarimage,gray2colornet,Color2Embed,coltrans,WACV} which can vividly colorize grayscale images. Though perform remarkably on images, these methods cannot be directly used to video colorization since they do not consider temporal coherency, producing flickering which severely affect subjective quality. To handle this issue, task-independent methods \cite{Blind,LearningBlind,TemporallyC,DeepPrior} utilize post-processing filter to smooth the generated images in temporal dimension. These methods work to some extent, but the generated color can be extremely different in adjacent frames, thus the results are still not continuous enough and tend to wash out colors. To further enhance temporal coherency, fully-automatic methods \cite{FullyAuto,automatic3D,AutomaticTC,VCGAN} are proposed. They directly embed the frame continuity in the model and map the grayscale frames to color frames. However, these methods are difficult to generate colorful result. Besides, in practical applications like old movie restoration, there are definite color in specific scenario for objects like clothes, skin, house, which have historical basis and are difficult to generate by fully-automatic approaches. 

The aforementioned problem can be better resolved by exemplar-based method with a certain reference image. Exemplar-based video colorization method utilize the colorized reference to lead the colorization of the whole grayscale video. Several CNNs (Convolutional Neural Networks) based methods \cite{VPN,trackingbycolor,Switchable,Deepremaster,DeepExamplar,referenceM} have been proposed for exemplar-based video colorization, and have achieved superior performance in still scenes or scenes with regular movement. But these methods always lack robustness in moving scenes, leading to color fading, color discontinuity or other artifacts when objects change greatly and irregularly. As shown in {\bf Fig. \ref{fig_1}(a)}, the dancer performs irregular movement, and recent state-of-the-art methods suffer from color fading and  generate discontinuous color. In {\bf Fig. \ref{fig_1}(b)}, the boy ride by a graffiti wall, as the texture on the wall changes, these methods suffer from severe color fading. We believe that this situation is mainly due to their lack of ability to model long-term dependencies. 

To improve the color robustness in moving scenes, this paper proposes an exemplar-based video colorization framework with long-term spatiotemporal dependency. (1) For long-term spatial dependency enhancement, a parallelized CNN-Transformer block and a double head non-local operation are designed. The CNN-Transformer block is more capable in modeling long-term spatial dependency, while keeping texture and structural features simultaneously \cite{conformer}. Besides, the proposed double head non-local operation further leverages the performance of augmented feature and improves the long-term spatial representation. (2) For long-term temporal dependency enhancement, we introduce the linkage subnet. The model computational complexity limits the length of input sequence. To handle long videos, we cut the frames into multiple frame blocks, and processes these frame blocks sequentially. Different from previous methods which mainly propagate motion information frame-by-frame \cite{Switchable,DeepExamplar,referenceM}, linkage subnet enables extra information propagation across adjacent frame blocks and maintains temporal continuity over a longer temporal range. Compared with the recent state-of-the-art methods, our method is able to generate more colorful and stabilized results, especially in moving scenes(Fig. \ref{fig_1}). Our contributions can be summarized as:

\begin{figure*}[!t]
  \centering
  
  \includegraphics[width=6.8in]{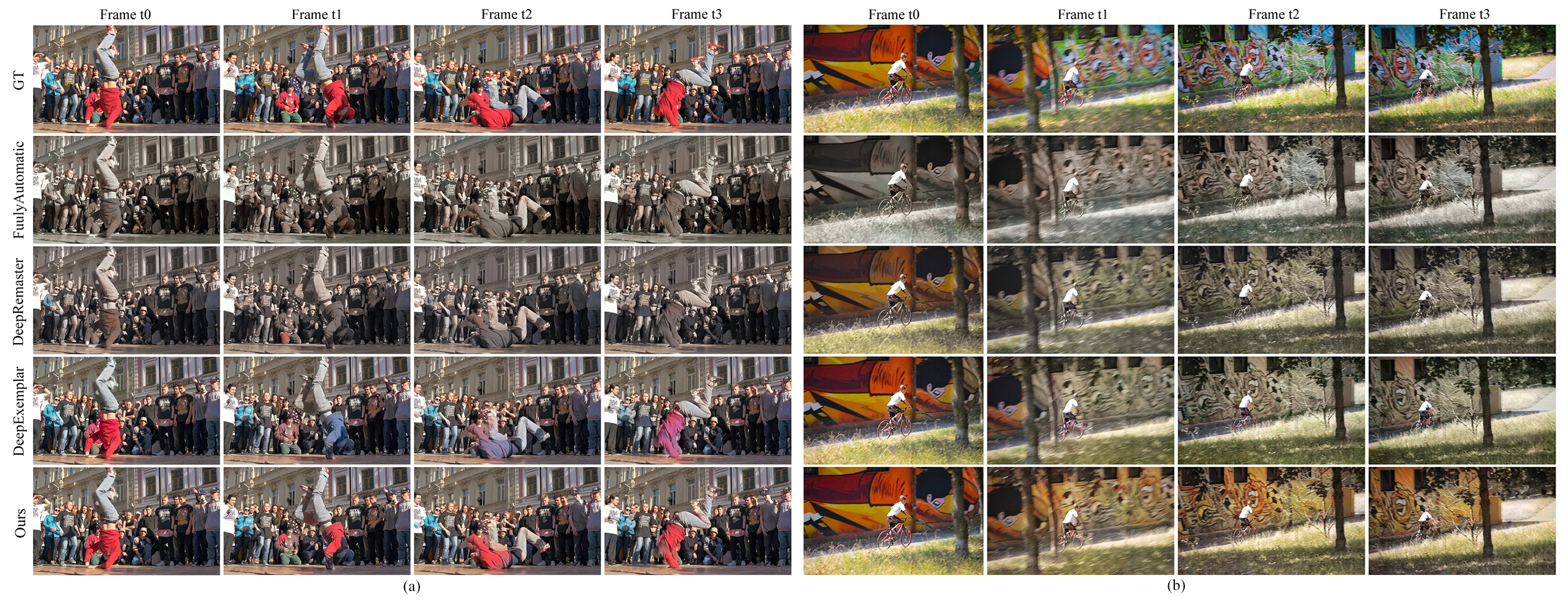} 
  
  \caption{Comparison with state-of-the-art video colorization methods on DAVIS \cite{davis} validation set. Each interval of two frames is greater than 10. In (a), the dancer performs irregular movement, while DeepRemaster \cite{Deepremaster} quickly suffers from color fading and DeepExemplar \cite{DeepExamplar} generates discontinuous color. In (b), the boy rides by a graffiti wall, as the texture on the wall changes, both DeepRemaster and DeepExemplar suffer from color fading. Differently, our method generate colorful and stabilized result. }
  \label{fig_1}
\end{figure*}

\begin{itemize}
  \item A novel framework for exemplar-based video colorization is proposed. Specially strengthened for robustness in moving scenes, our method has strong ability in modeling long-term dependency both spatially and temporally. 
  \item To enhance the long-term spatial dependency, we design a parallelized CNN-Transformer block and a double head non-local operation. The CNN-Transformer block can better incorporate long-term spatial dependency with local texture and structural features. The double head non-local operation further leverages the performance of augmented feature and improve the long-term spatial representation.
  \item  To enhance the long-term temporal dependency, we introduce linkage subnet. The linkage subnet keeps an information matrix of previous frame blocks and interact with present frame block via attention mechanism. It helps to retain temporal continuity between frame blocks in a video.
  \item Experiments demonstrate that our model outperforms recent state-of-the-art methods both quantitatively and qualitatively. And our model can generate more colorful, realistic and stabilized results, especially for scenes where objects change greatly and irregularly.
\end{itemize}

\section{Related Works}
In this section, we first introduce the three main methods in video colorization: task-independent, fully-automatic and exemplar-based, and then introduce the method of combining CNN with Transformer.
\subsection{Task-independent method}
Image colorization methods have made remarkable achievement. However, applying them directly to videos will produce severe temporal discontinuities. Task-independent methods \cite{Blind,LearningBlind,TemporallyC,DeepPrior} aim to correct the temporal discontinuities via post-processing. More precisely, they optimize a temporal filter by explicitly modeling and penalizing warp error computed by optical flows. Though these methods impressively alleviated flickering, the processed frames are still not continuous enough and tend to wash out colors. 

\subsection{Fully-automatic video colorization}
To further enhance temporal coherency, fully-automatic methods are proposed \cite{FullyAuto,automatic3D,AutomaticTC,VCGAN}. They directly map the grayscale frames to their color  embedding via deep neural networks, while considering frame continuity.  Lei et al. \cite{FullyAuto} propose a multimodal automatic framework which produce four diverse colorization results at a time. For preserving spatiotemporal consistency, they enforce similarity between pixel pairs, which are built by K nearest neighbor (KNN) search in feature space or following optical flows. Zhao et al. \cite{VCGAN} propose a hybrid recurrent network integrated both image and video colorization, together with a dense long-term loss which considers not only adjacent but long-term continuity. Nevertheless, these methods are difficult to generate colorful result. Besides, in practical applications like old movie restoration, there are definite color in specific scenario for objects like clothes, skin, house, which have historical basis and are difficult to generate by fully-automatic approaches.

\subsection{Exemplar-based video colorization}
Exemplar-based method utilizing one or more reference images as color guidance. Conventionally, the reference is generally a colored frame in video. These method leverage hand-craft low-level features to find temporal correspondence \cite{jacob2009colorization,ben2015approximate,xia2016robust}, and colorize following frames in sequence. While more recent methods tend to use deep neural networks to achieve temporal propagation \cite{VPN,trackingbycolor,Switchable}. These approaches generate much more colorful results, but their colorization only depends on the previous frame, thus they are easy to accumulate color errors in propagation. Another kind of methods refer to the reference images throughout the process \cite{Deepremaster,DeepExamplar,referenceM}, thus giving more stable results. For instance, Zhang et al. \cite{DeepExamplar} introduce a recurrent framework with novel loss functions, in which the colorization both depends on the reference and the previous frame. IIZUKA et al. \cite{Deepremaster} first propose a single framework for remastering vintage films. They adopt a source-reference attention that can handle multiple references, and utilize 3D-CNN for modeling temporal correspondence. Although favorable results are obtained, these approaches nonetheless lack long-term spatiotemporal dependencies, likely to wash out color in motion areas. Different from previous methods, our method has strong ability in modeling long-term dependency both spatially and temporally. 

\begin{figure*}[!t]
  \centering
  
  \includegraphics[width=6.8in]{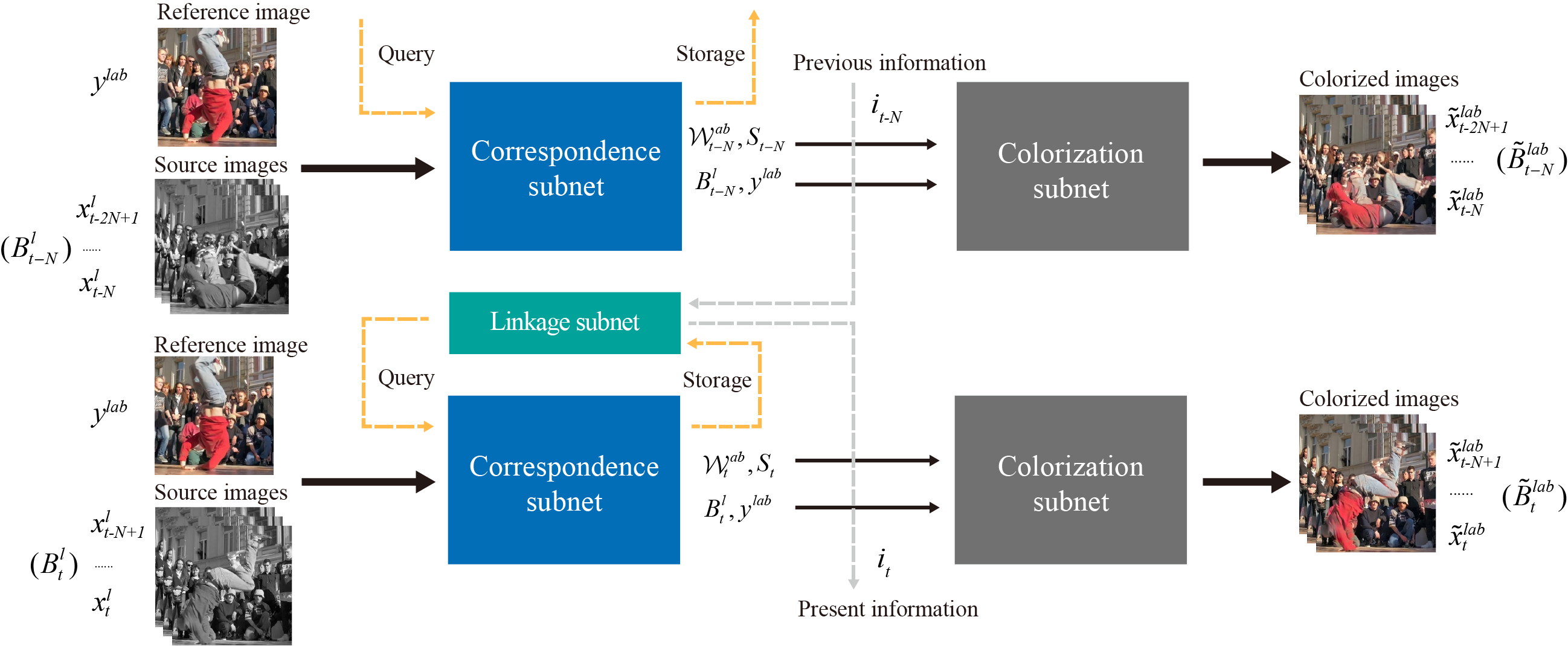}
  
  \caption{Overall framework of our approach.}
  \label{overall}
\end{figure*}

\subsection{Combining CNN with Transformer}
Though CNNs have strong performance on capturing local representation \cite{alexnet,krizhevsky2012imagenet,vgg,resnet,inception,densenet,deephighres}, they have difficulties in extracting global cues. One of the solution is to build deeper networks or employ more pooling layers to define larger reception field \cite{senet,genet}. Another is to include self-attention mechanism to augment local representations \cite{attentionaugment,relationnetwork}. But unfortunately, the first solution reduce spatial resolution which is important in colorization, while the second solution will deteriorate local feature details if attention is not properly integrated. 

Transformer \cite{attention} architecture is specialized in global feature correlation, and impressively advances computer vision tasks \cite{Vit,VTN,Vistr,coltrans}. However, it tends to ignore local feature details, which plays an important role in colorization. 

Recently, dual network structure is demonstrated effective in combining CNN-based local features with Transformer based global features in an interactive way \cite{conformer,mobileformer}. Standing on this viewpoint, we design a CNN-Transformer framework for augmenting feature representations in video colorization.

\section{Method}
\subsection{Overall framework}

The overall framework of our approach is depicted in {\bf Fig. \ref{overall}}. It mainly consists of three components: a Correspondence subnet $\mathcal N$, a Colorization subnet $\mathcal C$ and linkage subnet $\mathcal I$. The double head non-local operation is leveraged in the correspondence subnet and the CNN-Transformer is adopted in the correspondence subnet and colorization subnet.

Given a set of consistent grayscale video frames $X=\{x_1^l,x_2^l,x_3^l,...,x_L^l\}\in \mathbb{R}^{L \times 1 \times H \times W}$, our target is to generate a colorized video $\tilde{X}=\{\tilde{x}_1^{lab},\tilde{x}_2^{lab},\tilde{x}_3^{lab},...,\tilde{x}_L^{lab}\} \in \mathbb{R}^{L \times 3 \times H \times W}$ by a reference image $y^{lab}$, in which $l$ and $ab$ represent the luminance and chrominance in the CIELAB color space, and L represents the frame length. During processing, due to the memory limitation, we divide the entire video into multiple frame blocks and process them sequentially. Assume that the input block of $N$ frames at time $t$ denotes as $B_t^l=\{x_{t-N+1}^l,x_{t-N+2}^l,...,x_{t}^l\} \in \mathbb{R}^{N \times 1 \times H \times W}$. Firstly, the Correspondence subnet warp the reference color $y^{ab}$ onto frame block $B_t^l$ based on the correlation of their feature representations, obtain $\mathcal W^{ab} $ the warped color result and similarity map $S$ the correspondence reliability. Then utilizing $\mathcal W^{ab}$,$S$ and $y^{lab}$, the Colorization subnet colorizes $B_t^l$ to the final result $\tilde{B}_t^{lab}=\{\tilde{x}_{t-N+1}^{lab},\tilde{x}_{t-N+2}^{lab},...,\tilde{x}_{t}^{lab}\} \in \mathbb{R}^{N \times 3 \times H \times W}$. In order to generate temporal consistent frame sequences, linkage subnet is responsible to query information in previous frame blocks, while storage information of present frame block. In this way, we process video $X$ parallel and sequentially.

\subsection{Network architecture}

As mentioned before, our architecture mainly contains three components. In this subsection, we describe these subnets respectively.

\begin{figure*}[!t]
  \centering
  
  \includegraphics[width=6.8in]{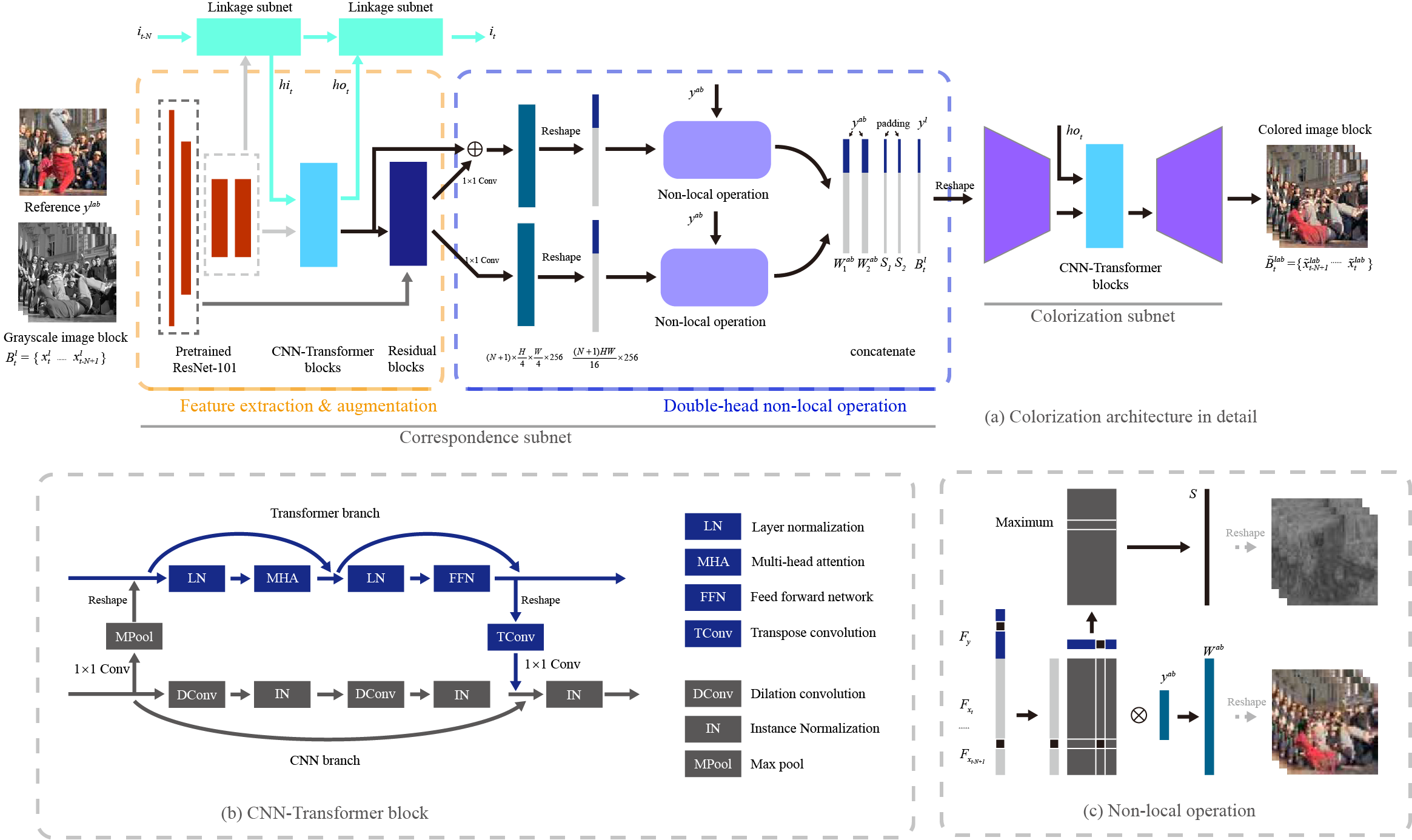}
  
  \caption{Illustrates our architecture in detail.}
  \label{detail}
  \hfil
\end{figure*}

{\bf Correspondence subnet}. The Correspondence subnet extracts and augments the features of the reference and grayscale images, then obtain coarse warped color results from colored reference. The detail architecture is illustrated in {\bf Fig. \ref{detail}(a)}.

Given the frame block $B_t^l$ and reference $y^{lab}$, firstly we need to build their feature representations. We start with extracting their feature sequences via a ResNet-101\cite{resnet} network pretrained on image classification, in which we remove the first maxpool layer to keep spatial resolution. Specifically, we take and concatenate the activation maps from layers of $relu1\_2$, $relu2\_3$ to form the low-level feature $R_{low}$, and $relu3\_3$, $relu3\_19$ to form the high-level feature $R_{high}$, which are found to work well in Experiment.

Moreover, For the high-level feature $R_{high}$, which contains more semantic information, we leverage CNN-Transformer blocks( as shown in {\bf Fig. \ref{detail}(b)}) to augment its spatiotemporal feature correlations. Concretely, $R_{high}$ is passed to the CNN branch, and $hi_t$ (containing previous frames' information, described below), is passed to the Transformer branch. Which obtain $R_{high}'$ (augmented high-level feature), and  $ho_t$ (containing both previous frames' and present frames' information), from the CNN and Transformer branch respectively. In CNN branch, coarse grained texture and structural features are extracted, and delivered to Transformer branch to supplement local representations. While in Transformer branch, feature correspondence is achieved in two aspects: (1) The correspondence among each pair of pixels in an image, rather than in the range of a convolution kernel, which provide the long-term dependency spatially; (2) The correspondence among semantically similar pixels in a block of different frames, which established temporal short-term dependency, and provide the ability to capture motion information. Similarly, the output of Transformer branch is delivered to CNN branch to supplement global representations. Then we combine augmented $R_{high}'$ with low-level feature $R_{low}$ acquiring the final representation $R'\in \mathbb{R}^{(N+1) \times 512 \times \frac{H}{4} \times \frac{H}{4}}$ ($N$ the frame block size and $1$ the reference) via residual blocks. During this process, the reference image is treated as the grayscale frames, and share the same network parameters. 

Finally, leveraging feature representation of $B_t^l$ and $y^{lab}$, we can find their dense correspondence by means of the double-head non-local operation. For feature $R' \in \mathbb{R}^{(N+1) \times 512 \times \frac{H}{4} \times \frac{H}{4}}$, we respectively contract its channel dimension by two $1 \times 1$ convolutions and flatten the resulting two features spatially and temporally to features $F_{k ,k \in \{ 1,2 \}}$ of size $\frac{(N+1) H W}{16} \times 256$. One of these features is directly passed to the non-local operation as a vanilla head, and the other is parametrically added by reshaped $R_{high}'$ to form an augmented head, which is explicitly modeled to contain more augmented spatiotemporal consistent features. Experimentally, we find that the augmented head is useful in hard assignments, but doesn't always perform well in common assignments. The double head non-local operation is proposed to combine the advantage of augmented head and vanilla head, further improves the robustness of spatial representation. 

The non-local operation is illustrated in {\bf Fig. \ref{detail}(c)}, which is basically the same as in \cite{non-local} except the parallel processing. The input feature $F \in \mathbb{R}^{\frac{(N+1) H W}{16} \times 256}$ is divided to $F_B \in \mathbb{R}^{\frac{N H W}{16} \times 256}$, $F_y \in \mathbb{R}^{\frac{H W}{16} \times 256}$ corresponding to feature representation of $B_t^l$ and $y^{lab}$, and by which the correlation matrix $\mathcal{M} \in \mathbb{R}^{ \frac{NHW}{16} \times \frac{HW}{16}}$ is computed. Each element $\mathcal{M}(i, j)$ measures the representation similarity of vector $F_{B}$ , $F_{y}$ at position $i$, $j$ respectively.
\begin{equation}
\begin{aligned}
  \label{deqn_ex1}
  \mathcal{M}_k(i, j)\!=\!\frac{\left(F_{B,k}(i)\!-\!\mu_{F_{B,k}}\right)\cdot\left(F_{y,k}(j)\!-\!\mu_{F_{y,k}}\right)}{\left\|F_{B,k}(i)\!-\!\mu_{F_{B,k}}\right\|_{2}\left\|F_{y,k}(j)\!-\!\mu_{F_{y,k}}\right\|_{2}} \ k\in \{1,2 \}
\end{aligned}
\end{equation}

$\mu_{F_{B}}$ and $\mu_{F_{y}}$ are mean feature vectors, and such normalization is empirically found stable in training \cite{DeepExamplar}. Utilizing the correlation matrix $\mathcal{M}$, we can warp the color from $y^{ab}$ to coarse warped result $\mathcal{W}^{a b}$. Specifically, using softmax operation to approach a one-hot vector along the row vector $\mathcal{M}(i, \cdot )$, then multiplied by $y^{ab}$ to sample the color in reference based on pixel-wise similarity, as formally described in {\bf Eq. (2)}.
\begin{equation}
  \label{deqn_ex1}
  \mathcal{W}^{a b}_k(i)=\sum_{j} \operatorname{softmax}(\mathcal{M}_k(i, j) / \tau) \cdot y^{a b}(j) \quad  k\in \{1,2 \}
\end{equation}
In which $\tau$ is a hyperparameter far less than 1. Furthermore, a similarity map is computed to mark the correspondence reliability of every sampled pixel.
\begin{equation}
  \label{deqn_ex1}
  \mathcal{S}_k(i)=\max _{j} \mathcal{M}_k(i, j) \quad k\in \{1,2 \}
\end{equation}
Each non-local head does the same process and totally obtain two pairs of outputs: $\mathcal{W}_1^{ab}$,$S_1$,$\mathcal{W}_2^{ab}$,$S_2$, which are then sent to the Colorization subnet as color guidance. The overall process in our Correspondence subnet can be formulated as:
\begin{equation}
  \label{deqn_ex1}
  (\mathcal{W}_1^{ab},S_1,\mathcal{W}_2^{ab},S_2,ho_t) = \mathcal N (B_t^l,y^{lab},hi_t)
\end{equation}

{\bf Colorization subnet}. The Colorization subnet is responsible for correcting the bad warped result with low similarity score, while generating color in faded region and smoothing the weird color. For better realizing spatiotemporal consistent colorization, we again incorporate CNN-Transformer blocks with encoder-decoder network. 

As shown in {\bf Fig. \ref{detail}}, the output of Correspondence subnet is concatenated with $B_t^{l}$,$y^{lab}$ to construct the input of CNN branch, and $ho_t$ is delivered to the Transformer branch. The colorization process can be simply summarized in {\bf Eq. (5)}, where $\oplus$ represents concatenating operation.
\begin{equation}
  \label{deqn_ex1}
  \tilde B_t^{ab} = \mathcal C (\mathcal{W}_1^{ab} \oplus S_1 \oplus \mathcal{W}_2^{ab} \oplus S_2 \oplus B_t^l \oplus y^{lab},ho_t)
\end{equation}

With input luminance channel $B_t^l$, the colorization result at time $t$ is indicated as $\tilde B_t^{lab}$ or $\tilde B_t = \{\tilde{x}_{t-N+1},\tilde{x}_{t-N+2},...,\tilde{x}_{t}\}$.

{\bf Linkage subnet}. While frames in the same block interact via CNN-Transformer blocks, linkage subnet interact the information across frame blocks along the entire video. We divide this task into two parts: query and storage.

Depending on high-level representation of input frame block $R_{high}$, we request the hidden state $hi_t$ from previous information $i_{t-N}$ base on their representation similarity, acquiring related information from previous frames. This query process can be modeled by an attention mechanism:
\begin{equation}
  \label{deqn_ex1}
    \left\{
                 \begin{array}{lr}
                 hi_t = R_{high} + \operatorname{softmax} (\frac{R_{high},i_{t-N}^T}{\sqrt{d}})i_{t-N}, & t>0 \\
                 hi_t = R_{high}, &  t=0
                 \end{array}
    \right.
\end{equation}

$d$ is the scale factor for normalization, note that the reshape process is omitted for brevity. Together with residual linear layers, the linkage subnet for query is constructed. Identically, the storage process reserve hidden state $ho_t$, containing present information to construct updated $i_t$, which is then delivered to the next recurrence:
\begin{equation}
  \label{deqn_ex1}
  \left\{
               \begin{array}{lr}
               i_t = i_{t-N} + \operatorname{softmax} (\frac{i_{t-N},ho_t^T}{\sqrt{d}})ho_t, & t>0 \\
               i_t = ho_t, &  t=0
               \end{array}
  \right.
  \end{equation}
\subsection{Loss}
Our network aims to colorize grayscale video frames via a colored reference image. The colorization result is supposed to be colorful and realistic, especially in motion areas, while retain temporal consistency. To this end, we employ the following objective losses:

{\bf L1 loss}. In order to determine pixel level colorization accuracy, we adopt L1 loss which computes pixel difference between chrominance of output $\tilde x_t$ and the ground truth $x_t$, which is demonstrated to generate more distinct color than L2 loss\cite{long2016learning,mathieu2015deep,niklaus2017video}. The standard L1 loss can be written as:
\begin{equation}
  \label{deqn_ex1}
  \mathcal L_{L1} = \|   \tilde x_t^{ab} - x_t^{ab} \|_1
\end{equation}
{\bf Perceptual loss}. Besides pixel level accuracy, we expect the colorization result perceptually similar to the ground truth. The perceptual loss measures semantic distance by comparing high-level features extracted in pretrained VGG19\cite{vgg} network. 
\begin{equation}
  \label{deqn_ex1}
  \mathcal L_{perc} = \| \Phi_L(\tilde x) - \Phi_L(x) \|_2^2
\end{equation}
Where $\Phi_L$ represents activation map extracted at $L$-th layer. We set $ L = relu5\_2$ since the top layers contain more semantic information.

{\bf Temporal loss}. Generated colors should be consistent along optical trajectory in videos, thus the temporal loss is employed to measure the warp error in temporal dimension. 
\begin{equation}
  \label{deqn_ex1}
  \mathcal L_{temp} = \| M_{t-1,t}\odot W_{t-1,t}(\tilde x_{t-1}^{ab}) - M_{t-1,t}\odot \tilde x_{t}^{ab} \|_1
\end{equation}
Where $W_{t-1,t}$ and $M_{t-1,t}$ are forward optical flow and occlusion mask from frame $x_{t-1}$ to $x_t$.

In addition, we also adopt adversarial loss and smooth loss introduced in \cite{DeepExamplar}, which help to generate vivid image and penalize color bleeding. In conclusion, our total objective loss can be written as:
\begin{equation}
  \begin{aligned}
  \mathcal L_{total} & = \lambda_{L1} \mathcal L_{L1} + \lambda_{perc} \mathcal L_{perc} + \lambda_{temp} \mathcal L_{temp} \\ & + \lambda_{adv} \mathcal L_{adv} + \lambda_{smooth} \mathcal L_{smooth}
 \end{aligned}
 \label{deqn_ex1}
\end{equation}

\begin{table*}[]
  \centering
  \caption{Quantitative comparison with state-of-the-art video colorization methods on DAVIS2017 and Videvo dataset. The number in red represents the best and blue the second-best result. The original network of DeepRemaster \cite{Deepremaster} contains a denoising subnet which is irrelevant to the test datasets used in this paper, we remove this subnet and represent the rest of the network via '$*$'.}
  \label{tab:comparision}
\resizebox{\textwidth}{!}{
  \begin{tabular}{c|ccccc|ccccc|c}
  \hline
  {\color[HTML]{000000} } &
    \multicolumn{5}{c|}{{\color[HTML]{000000} DAVIS}} &
    \multicolumn{5}{c|}{{\color[HTML]{000000} Videvo}} &
    {\color[HTML]{000000} } \\
  {\color[HTML]{000000} } &
    {\color[HTML]{000000} FID$\downarrow$} &
    {\color[HTML]{000000} LPIPS$\downarrow$} &
    {\color[HTML]{000000} PSNR$\uparrow$} &
    {\color[HTML]{000000} COLOR$\uparrow$} &
    {\color[HTML]{000000} WE$\downarrow$} &
    {\color[HTML]{000000} FID$\downarrow$} &
    {\color[HTML]{000000} LPIPS$\downarrow$} &
    {\color[HTML]{000000} PSNR$\uparrow$} &
    {\color[HTML]{000000} COLOR$\uparrow$} &
    {\color[HTML]{000000} WE$\downarrow$} &
    {\color[HTML]{000000} Method} \\ \hline
    {\color[HTML]{000000} ChromaGAN \cite{Chromagan}} &
    {\color[HTML]{000000} 147.11} &
    {\color[HTML]{000000} 0.1528} &
    {\color[HTML]{000000} 23.57} &
    {\color[HTML]{000000} 12.58} &
    {\color[HTML]{000000} 8.94} &
    {\color[HTML]{000000} 142.49} &
    {\color[HTML]{000000} 0.1637} &
    {\color[HTML]{000000} 23.81} &
    {\color[HTML]{4472C4} 12.77} &
    {\color[HTML]{000000} 6.760} &
    {\color[HTML]{000000} Image (Fully-automatic)} \\
  {\color[HTML]{000000} Gray2ColorNet \cite{gray2colornet}} &
    {\color[HTML]{000000} 92.02} &
    {\color[HTML]{000000} 0.1536} &
    {\color[HTML]{000000} 23.23} &
    {\color[HTML]{FF0000} 14.73} &
    {\color[HTML]{000000} 3.759} &
    {\color[HTML]{000000} 100.10} &
    {\color[HTML]{000000} 0.1445} &
    {\color[HTML]{000000} 25.21} &
    {\color[HTML]{FF0000} 12.79} &
    {\color[HTML]{000000} 2.017} &
    {\color[HTML]{000000} Image (Exemplar-based)} \\
    {\color[HTML]{000000} Color2Embed \cite{Color2Embed}} &
    {\color[HTML]{000000} 89.19} &
    {\color[HTML]{000000} 0.0858} &
    {\color[HTML]{000000} 27.64} &
    {\color[HTML]{000000} 12.67} &
    {\color[HTML]{000000} 2.520} &
    {\color[HTML]{000000} 95.78} &
    {\color[HTML]{000000} 0.1010} &
    {\color[HTML]{000000} 27.32} &
    {\color[HTML]{000000} 10.81} &
    {\color[HTML]{000000} 1.860} &
    {\color[HTML]{000000} Image (Exemplar-based)} \\
    {\color[HTML]{000000} ChromaGAN+DVP \cite{DeepPrior}} &
    {\color[HTML]{000000} 147.59} &
    {\color[HTML]{000000} 0.1420} &
    {\color[HTML]{000000} 23.98} &
    {\color[HTML]{000000} 11.22} &
    {\color[HTML]{000000} 2.138} &
    {\color[HTML]{000000} 139.99} &
    {\color[HTML]{000000} 0.1493} &
    {\color[HTML]{000000} 24.15} &
    {\color[HTML]{000000} 12.15} &
    {\color[HTML]{000000} 1.922} &
    {\color[HTML]{000000} Task-independent} \\
    {\color[HTML]{000000} Gray2ColorNet+DVP \cite{DeepPrior}} &
    {\color[HTML]{000000} 96.98} &
    {\color[HTML]{000000} 0.1151} &
    {\color[HTML]{000000} 26.54} &
    {\color[HTML]{000000} 13.77} &
    {\color[HTML]{000000} 2.565} &
    {\color[HTML]{000000} 104.65} &
    {\color[HTML]{000000} 0.1285} &
    {\color[HTML]{000000} 26.20} &
    {\color[HTML]{000000} 11.37} &
    {\color[HTML]{000000} 1.362} &
    {\color[HTML]{000000} Task-independent} \\
    {\color[HTML]{000000} Color2Embed+DVP \cite{DeepPrior}} &
    {\color[HTML]{000000} 89.19} &
    {\color[HTML]{000000} 0.0866} &
    {\color[HTML]{000000} 27.56} &
    {\color[HTML]{000000} 12.21} &
    {\color[HTML]{000000} 1.837} &
    {\color[HTML]{000000} 98.54} &
    {\color[HTML]{000000} 0.1010} &
    {\color[HTML]{000000} 27.4} &
    {\color[HTML]{000000} 10.16} &
    {\color[HTML]{000000} 1.04} &
    {\color[HTML]{000000} Task-independent} \\
  {\color[HTML]{000000} FullyAutomatic \cite{FullyAuto}} &
    {\color[HTML]{000000} 168.47} &
    {\color[HTML]{000000} 0.1571} &
    {\color[HTML]{000000} 23.94} &
    {\color[HTML]{000000} 7.23} &
    {\color[HTML]{FF0000} 0.944} &
    {\color[HTML]{000000} 137.28} &
    {\color[HTML]{000000} 0.1459} &
    {\color[HTML]{000000} 25.23} &
    {\color[HTML]{000000} 6.36} &
    {\color[HTML]{000000} 0.629} &
    {\color[HTML]{000000} Fully-automatic} \\
  {\color[HTML]{000000} DeepRemaster \cite{Deepremaster}} &
    {\color[HTML]{000000} 118.79} &
    {\color[HTML]{000000} 0.1268} &
    {\color[HTML]{000000} 26.47} &
    {\color[HTML]{000000} 9.70} &
    {\color[HTML]{000000} 1.575} &
    {\color[HTML]{000000} 117.80} &
    {\color[HTML]{000000} 0.1365} &
    {\color[HTML]{000000} 26.98} &
    {\color[HTML]{000000} 7.48} &
    {\color[HTML]{4472C4} 0.626} &
    {\color[HTML]{000000} Exemplar-based} \\
  {\color[HTML]{000000} DeepRemaster* \cite{Deepremaster}} &
    {\color[HTML]{000000} 104.63} &
    {\color[HTML]{000000} 0.0971} &
    {\color[HTML]{000000} 26.22} &
    {\color[HTML]{000000} 9.48} &
    {\color[HTML]{4472C4} 1.446} &
    {\color[HTML]{000000} 110.26} &
    {\color[HTML]{000000} 0.1083} &
    {\color[HTML]{000000} 26.61} &
    {\color[HTML]{000000} 7.43} &
    {\color[HTML]{FF0000} 0.597} &
    {\color[HTML]{000000} Exemplar-based} \\
  {\color[HTML]{000000} DeepExemplar \cite{DeepExamplar}} &
    {\color[HTML]{4472C4} 57.50} &
    {\color[HTML]{4472C4} 0.0687} &
    {\color[HTML]{4472C4} 31.05} &
    {\color[HTML]{000000} 14.20} &
    {\color[HTML]{000000} 1.931} &
    {\color[HTML]{4472C4} 65.28} &
    {\color[HTML]{4472C4} 0.0808} &
    {\color[HTML]{4472C4} 29.96} &
    {\color[HTML]{000000} 12.51} &
    {\color[HTML]{000000} 1.294} &
    {\color[HTML]{000000} Exemplar-based} \\
  {\color[HTML]{000000} ours} &
    {\color[HTML]{FF0000} 54.52} &
    {\color[HTML]{FF0000} 0.0621} &
    {\color[HTML]{FF0000} 31.77} &
    {\color[HTML]{4472C4} 14.37} &
    {\color[HTML]{000000} 2.486} &
    {\color[HTML]{FF0000} 59.77} &
    {\color[HTML]{FF0000} 0.0770} &
    {\color[HTML]{FF0000} 30.50} &
    {\color[HTML]{000000} 11.66} &
    {\color[HTML]{000000} 1.599} &
    {\color[HTML]{000000} Exemplar-based} \\ \hline
  \end{tabular}
}
\end{table*}

\section{Implementation}
{\bf Network Structure}. All the Transformers in our framework employ Layer Normalization, 3D sine position embedding proposed in \cite{Vistr},and the Feed Forward Network (FFN) consists of 2 linear layers in residual form, while the head number in Multi-Head Attention (MHA) is 6 for Correspondence subnet and 4 for others. The CNN blocks employ Instance Normalization, and consist of 2 to 3 convolutional layers. Besides, the number of Residual blocks and CNN-Transformer blocks are both set to 3. For the encoder-decoder in Colorization subnet, 3 encoder blocks and decoder blocks together with skip connection are adopted, and the decoder ends with a tanh activation function to constraint output color range. 

{\bf Datasets.} In training, we involve three video datasets: DAVIS\cite{davis}, FVI\cite{FVI} and Videvo \cite{videvo}. DAVIS is a widely used VOS (Video Object Segmentation) dataset, including 60/30 training/testing videos, each contains about 100 frames. Follow the setting of \cite{zou2021progressive}, we use FVI sampled from the Youtube-VOS\cite{YoutubeVOS} dataset, containing 1940/100 training/testing videos and there are about 40 frames in a video clip. Videvo\cite{videvo} mostly contains long videos of about 300 frames, having 90/20 for training/testing. Totally, we get 2090 training videos. And in testing, we use the DAVIS and Videvo datasets (test respectively), which are widely used for comparison in video colorization tasks \cite{DeepExamplar,VCGAN,referenceM,FullyAuto}. 

{\bf Training.}  In training, we randomly select a video clips with max length of 20 frames, and use the first frame as reference. Optical flows are generated by RAFT\cite{raft}, and the occlusion masks are computed follow the method in \cite{ruder2016artistic}. All the selected frames are resized to 384 $\times$ 216 to suit the standard aspect ratio 16:9. 

Since video segmentation shows close relation to video colorization \cite{Switchable,trackingbycolor}, we utilize the pretrained video instance segmentation network Vistr\cite{Vistr} to initialize the Transformer branch in CNN-Transformer blocks. The hyperparameters are set as: $\gamma=0.01$, $\alpha = 0.75$,$N=2$, $\lambda_{L1}=0.5$,$\lambda_{perc}=0.05$,$\lambda_{temp}=3.0$,$\lambda_{adv}=0.2$ and $\lambda_{smooth}=4.0$. We train the network for 10 epochs using the AdamW optimizer with parameters $\beta_1=0.5$,$\beta_2=0.999$, weight decay $1 \times 10^{-4}$, and the learning rate is set to $1 \times 10^{-3}$,$1 \times 10^{-5}$,$1 \times 10^{-4}$ for discriminator, pretrained Resnet-101 network and others, decayed by 10, 100 at epoch 4, 8.

All the experiments are implemented on a single NVIDIA 3090 GPU.

\section{Experiment}
In this section, we compare our methods with various state-of-the-art approaches. Then, we investigate the effectiveness of each proposed component. After that, some further analysis on the double non-local heads is given, by visualizing warped color results of the double non-local heads. Besides, we also make parameter analysis on the selection of frame number N in a frame block processed in parallel. Finally, we discuss the limitations of this method, and illustrate some failure cases.

\begin{figure*}[t]
  \centering
  
  \includegraphics[width=6.8in]{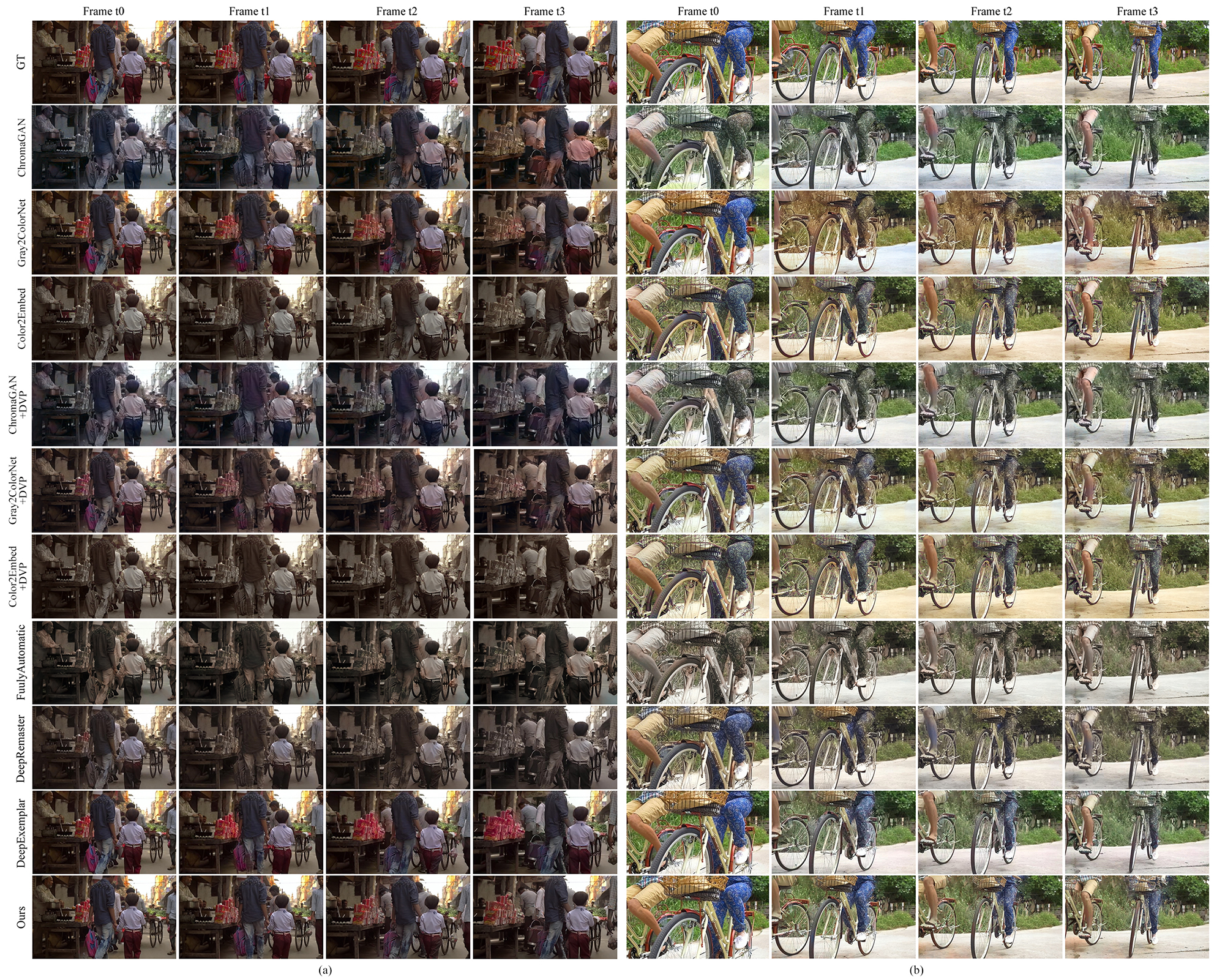}
  
  \caption{Colorization results on Videvo \cite{videvo} test set. The average interval of two frames is greater than 10. In {\bf (a)}, as the man walks along the alley, the goods on the table get closer and the backpack is rotated. Only Gray2ColorNet \cite{gray2colornet}, DeepExemplar \cite{DeepExamplar} and our method retain colorful results. However, Gray2ColorNet fail in frame t3 (the color on the goods is faded) and DeepExemplar suffers color bleeding and discontinuity. In contrast, our method generates more distinct and continuous color. In {\bf (b)}, the couple ride bicycle on the road. The state-of-the-art methods quickly suffer color fading especially on the blue trousers, while our results are more colorful and realistic.}
  \label{videvo}
\end{figure*}

\begin{figure*}[t]
  \centering
  
  \includegraphics[width=6.8in]{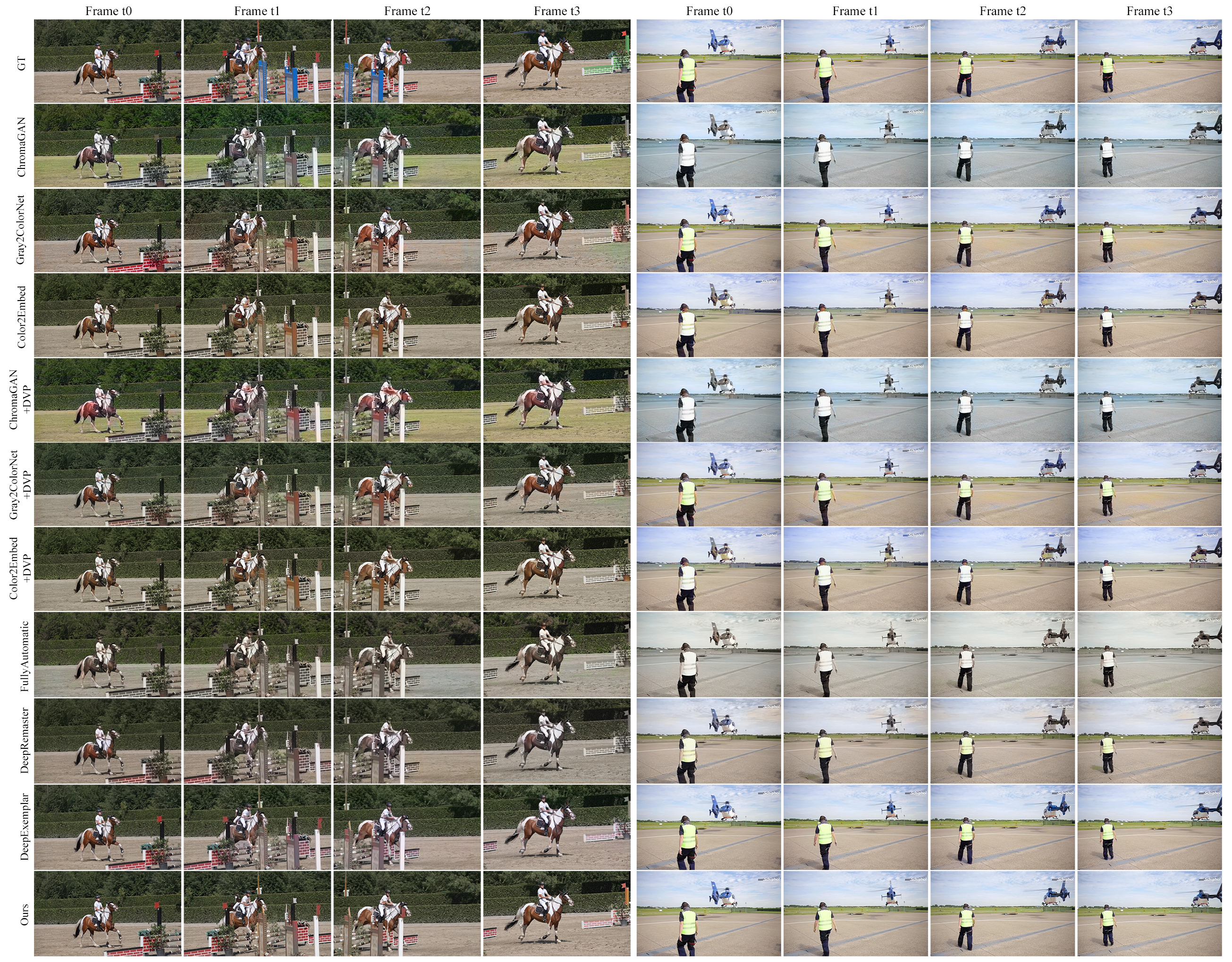}
  
  \caption{Colorization results on DAVIS \cite{davis} test set. Each interval of two frames is greater than 10. In (a), the horse jumps over the sticks and are partially occluded in frame t1, t2. Missing temporal information, image methods suffer from color discontinuity and hard to generate distinct color. With DVP \cite{DeepPrior}, the color continuity improves a lot, but the results are not colorful enough. While our method gives plausible result. In (b), as the man walks over and the helicopter rotates, Gray2ColorNet \cite{gray2colornet}, DeepExemplar \cite{DeepExamplar} and our method have comparable colorful results. However, Gray2ColorNet has blue artifact on the ground and DeepExemplar has red artifact on the helicopter in frame t3. Both the analyses above demonstrate that our method gives more stabilized and realistic color in moving scenes or objects.} 
  \label{davis}
\end{figure*}

\subsection{Comparisons with state-of-the-arts}
We compare our approach against various video colorization methods both quantitatively and qualitatively. The baseline include exemplar-based methods (DeepExemplar \cite{DeepExamplar}, DeepRemaster \cite{Deepremaster}), image colorization methods (ChromaGAN \cite{Chromagan}, Gray2ColorNet \cite{gray2colornet} and Color2Embed \cite{Color2Embed}) with task-independent method (DVP \cite{DeepPrior}), and fully-automatic method (FullyAutomatic \cite{FullyAuto}), which are regarded as state-of-the-art.

{\bf Quantitative comparison}. The performances are evaluated on 5 metrics: image metrics FID (Fr´echet Inception Distance) \cite{fid}, LPIPS (Learned Perceptual Image Patch Similarity) \cite{lpips}, PSNR (Peak Signal to Noise Ratio), COLOR (colorfulness metric) \cite{color} and video metric WE (Warp Error) \cite{LearningBlind}. The results are reported in {\bf Table \ref{tab:comparision}}. First, FID and LPIPS  are utilized to measure the semantic distance between generated results and realistic images. Our method achieves the best FID and LPIPs, which represents our network generates most perceptually realistic results. Second, PSNR is utilized to evaluate the pixel-wise colorization accuracy. Our method obtains the highest PSNR, which indicates our network generates most accurate color compared with the ground truth. Besides, we evaluate the methods on the COLOR metric, which measures image color richness. Gray2ColorNet \cite{gray2colornet} gives the best COLOR, while Gray2ColorNet+DVP \cite{DeepPrior}, ChromaGAN \cite{Chromagan},DeepExemplar \cite{DeepExamplar} and our method gives comparable results. In addition, we take the WE to measure the temporal consistency of the output videos. As illustrated, FullyAutomatic \cite{FullyAuto} and DeepRemaster \cite{Deepremaster} achieve lowest WE, but we claim that it is due to their faded color (As shown in {\bf Fig. 1}) to a certain extent. As color faded, the differences between adjacent frames decline, which is also validated on the other metrics, such as their low COLOR scores. In summary, our approach obtain the best generation quality.

{\bf Qualitative comparison}. {\bf Fig. \ref{videvo}} visualize the colorization results of each method on Videvo test set. In {\bf Fig. \ref{videvo}(a)}, as the man walks along the alley, the goods on the table get closer and the backpack is rotated. Only Gray2ColorNet \cite{gray2colornet}, DeepExemplar \cite{DeepExamplar} and our method retain colorful results. However, Gray2ColorNet fails in frame t3 (the color on the goods is faded) and DeepExemplar suffers color bleeding (the goods) and discontinuity (the backpack). In contrast, our method generates more distinct and continuous color. In {\bf Fig. \ref{videvo}(b)}, the couple ride bicycle on the road. The state-of-the-art methods quickly suffer color fading especially on the blue trousers, while our results are more colorful and realistic. 

In addition, we compare the methods visually on DAVIS test set. In {\bf Fig. \ref{davis}(a)}, the horse jumps over the sticks and are partially occluded in frame t1, t2. Missing temporal information, image colorization methods suffer from color discontinuity and are hard to generate distinct color, especially on the horse and the brick wall. With DVP, the color continuity improves a lot, but the result images are still unrealistic. While our method generates more plausible result. In {\bf Fig. \ref{davis}(b)}, as the man walks over and the helicopter rotates, Gray2ColorNet, DeepExemplar and our method have comparable colorful results. However, Gray2ColorNet has blue artifact on the ground and DeepExemplar has red artifact on the helicopter in frame t3. Both the analyses above demonstrate that our method gives more stabilized and realistic color in moving scenes or objects.

\begin{table*}[!t]
  \centering
  \caption{Quantitative comparison without individual components proposed in this paper. The number in red represents the best and blue the second-best result. Our full architecture has the most comprehensive performance.}
  \label{tab:ablation}
\resizebox{\textwidth}{!}{
  \begin{tabular}{c|ccccc|ccccc}
  \hline
   &
    \multicolumn{5}{c|}{DAVIS} &
    \multicolumn{5}{c}{Videvo} \\
   &
   {\color[HTML]{000000} FID $\downarrow$} &
   {\color[HTML]{000000} LPIPS $\downarrow$} &
   {\color[HTML]{000000} PSNR  $\uparrow$} &
   {\color[HTML]{000000} COLOR $\uparrow$} &
   {\color[HTML]{000000} WE $\downarrow$} &
   {\color[HTML]{000000} FID $\downarrow$} &
   {\color[HTML]{000000} LPIPS $\downarrow$} &
   {\color[HTML]{000000} PSNR  $\uparrow$} &
   {\color[HTML]{000000} COLOR $\uparrow$} &
   {\color[HTML]{000000} WE $\downarrow$} \\ \hline
   \begin{tabular}[c]{@{}c@{}}w/o Transformer Branch \& Linkage Subnet \end{tabular} &
    {\color[HTML]{FF0000} 54.10} &
    {\color[HTML]{000000} 0.0632} &
    {\color[HTML]{000000} 31.39} &
    {\color[HTML]{000000} 14.18} &
    {\color[HTML]{000000} 2.684} &
    {\color[HTML]{000000} 60.90} &
    {\color[HTML]{000000} 0.0795} &
    {\color[HTML]{000000} 30.18} &
    {\color[HTML]{000000} 11.41} &
    {\color[HTML]{000000} 1.815} \\
  w/o Double-head &
    {\color[HTML]{000000} 56.12} &
    {\color[HTML]{000000} 0.0651} &
    {\color[HTML]{000000} 31.38} &
    {\color[HTML]{000000} 14.24} &
    {\color[HTML]{4472C4} 2.495} &
    {\color[HTML]{000000} 60.73} &
    {\color[HTML]{000000} 0.0798} &
    {\color[HTML]{000000} 30.16} &
    {\color[HTML]{000000} 11.33} &
    {\color[HTML]{FF0000} 1.504} \\
    w/o Linkage Subnet &
    {\color[HTML]{000000} 54.55} &
    {\color[HTML]{FF0000} 0.0620} &
    {\color[HTML]{FF0000} 31.77} &
    {\color[HTML]{FF0000} 14.46} &
    {\color[HTML]{000000} 2.537} &
    {\color[HTML]{4472C4} 59.79} &
    {\color[HTML]{4472C4} 0.0771} &
    {\color[HTML]{FF0000} 30.50} &
    {\color[HTML]{FF0000} 11.74} &
    {\color[HTML]{000000} 1.618} \\
  Full Architecture &
    {\color[HTML]{4472C4} 54.52} &
    {\color[HTML]{4472C4} 0.0621} &
    {\color[HTML]{FF0000} 31.77} &
    {\color[HTML]{4472C4} 14.37} &
    {\color[HTML]{FF0000} 2.486} &
    {\color[HTML]{FF0000} 59.77} &
    {\color[HTML]{FE0000} 0.0770} &
    {\color[HTML]{FE0000} 30.50} &
    {\color[HTML]{4472C4} 11.65} &
    {\color[HTML]{4472C4} 1.599} \\ \hline
  \end{tabular}
}
\end{table*}

\subsection{Ablation studies}  
To evaluate the effect of the Transformer channel in CNN-Transformer module, the augmented head in double-head non-local operation, and  linkage subnet proposed in this paper, we remove the individual network components respectively, and quantitative comparison is shown in {\bf Table \ref{tab:ablation}}. Additionally, some further analysis on the double non-local heads is given. Besides, we also make parameter analysis on the selection of frame number N in a frame block processed in parallel.

{\bf Effect of CNN-Transformer block.} Since linkage subnet relies on the output of Transformer branch of CNN-Transformer block, we remove the Transformer branch and the linkage subnet together. Without Transformer branch and linkage subnet, both the image quality (FID, LPIPS, PSNR, COLOR) and video quality (WE) decline, because the long-term spatial and temporal modelling capabilities are weakened.

{\bf Effect of double-head non-local operation.} Without augmented head, all the image colorization metrics decline and the WE improved contradictorily. This is because in double-head non-local operation, the augmented head contains more long-term spatial representation. It reveals that the augmented features helps to generate more vivid and accurate image results.

\begin{figure}[t]
  \centering
  
  \includegraphics[width=3.2in]{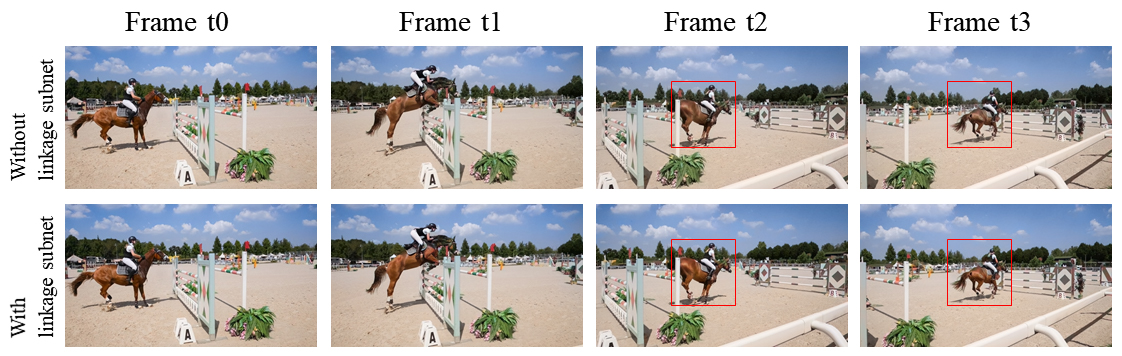}
  
  \caption{An example of colorization with or without linkage subnet. Without linkage subnet, the video's color tend to be faded on the moving object (the horse). While with linkage subnet, the video can preserve consistent color. This represents the capability of linkage subnet to enhance long-term temporal dependency}
  \label{wo_linkage}
\end{figure}

{\bf Effect of linkage subnet.} Without linkage subnet, the image quality remain but get higher WE. This proves the ability of linkage subnet to maintain video coherency. One may notice that the image metrics COLOR improves with video metric WE worsens. We assume that there is a certain contradiction between them. Besides, an example of colorization with or without linkage subnet is illustrated in {\bf Fig. \ref{wo_linkage}}. Without linkage subnet, the video's color tend to be faded on the moving object (the horse). While with linkage subnet, the video can preserve consistent color. This represents the capability of linkage subnet to enhance long-term temporal dependency.

\begin{table*}[!t]
  \caption{Quantitative comparison with different number of $N$, the best performance is achieved at $N=3$ and 5 respectively for DAVIS and Videvo datasets, which are the most efficient while remain stable quality.}
  \label{tab:frames}
  \resizebox{\textwidth}{!}{
  \begin{tabular}{c|cccccc|cccccc}
  \hline
  {\color[HTML]{000000} } &
    \multicolumn{6}{c|}{{\color[HTML]{000000} DAVIS}} &
    \multicolumn{6}{c}{{\color[HTML]{000000} Videvo}} \\
  {\color[HTML]{000000} N} &
  {\color[HTML]{000000} FID $\downarrow$} &
  {\color[HTML]{000000} LPIPS $\downarrow$} &
  {\color[HTML]{000000} PSNR  $\uparrow$} &
  {\color[HTML]{000000} COLOR $\uparrow$} &
  {\color[HTML]{000000} WE $\downarrow$} &
  {\color[HTML]{000000} DUR $\downarrow$} &
  {\color[HTML]{000000} FID $\downarrow$} &
  {\color[HTML]{000000} LPIPS $\downarrow$} &
  {\color[HTML]{000000} PSNR  $\uparrow$} &
  {\color[HTML]{000000} COLOR $\uparrow$} &
  {\color[HTML]{000000} WE $\downarrow$} &
  {\color[HTML]{000000} DUR $\downarrow$} \\ \hline
  {\color[HTML]{000000} 1} &
    {\color[HTML]{000000} 54.64} &
    {\color[HTML]{000000} 0.0621} &
    {\color[HTML]{000000} 31.77} &
    {\color[HTML]{000000} 14.35} &
    {\color[HTML]{000000} 2.485} &
    {\color[HTML]{000000} 0.1001} &
    {\color[HTML]{000000} 59.71} &
    {\color[HTML]{000000} 0.0769} &
    {\color[HTML]{000000} 30.51} &
    {\color[HTML]{000000} 11.66} &
    {\color[HTML]{000000} 1.600} &
    0.0993 \\
  {\color[HTML]{000000} 2} &
    {\color[HTML]{000000} 54.54} &
    {\color[HTML]{000000} 0.0621} &
    {\color[HTML]{000000} 31.77} &
    {\color[HTML]{000000} 14.37} &
    {\color[HTML]{000000} 2.485} &
    {\color[HTML]{000000} 0.0743} &
    {\color[HTML]{000000} 59.79} &
    {\color[HTML]{000000} 0.0770} &
    {\color[HTML]{000000} 30.51} &
    {\color[HTML]{000000} 11.66} &
    {\color[HTML]{000000} 1.600} &
    0.0704 \\
  {\color[HTML]{000000} 3} &
    {\color[HTML]{000000} 54.52} &
    {\color[HTML]{000000} 0.0621} &
    {\color[HTML]{000000} 31.77} &
    {\color[HTML]{000000} 14.37} &
    {\color[HTML]{000000} 2.486} &
    {\color[HTML]{FF0000} 0.0706} &
    {\color[HTML]{000000} 59.77} &
    {\color[HTML]{000000} 0.0770} &
    {\color[HTML]{000000} 30.50} &
    {\color[HTML]{000000} 11.65} &
    {\color[HTML]{000000} 1.599} &
    {\color[HTML]{000000} 0.0630} \\
  {\color[HTML]{000000} 4} &
    {\color[HTML]{000000} 54.65} &
    {\color[HTML]{000000} 0.0621} &
    {\color[HTML]{000000} 31.77} &
    {\color[HTML]{000000} 14.38} &
    {\color[HTML]{000000} 2.487} &
    {\color[HTML]{000000} 0.0730} &
    {\color[HTML]{000000} 59.74} &
    {\color[HTML]{000000} 0.0770} &
    {\color[HTML]{000000} 30.50} &
    {\color[HTML]{000000} 11.66} &
    {\color[HTML]{000000} 1.600} &
    0.0630 \\
  {\color[HTML]{000000} 5} &
    {\color[HTML]{000000} 54.63} &
    {\color[HTML]{000000} 0.0621} &
    {\color[HTML]{000000} 31.77} &
    {\color[HTML]{000000} 14.38} &
    {\color[HTML]{000000} 2.492} &
    {\color[HTML]{000000} 0.0774} &
    {\color[HTML]{000000} 59.77} &
    {\color[HTML]{000000} 0.0770} &
    {\color[HTML]{000000} 30.50} &
    {\color[HTML]{000000} 11.67} &
    {\color[HTML]{000000} 1.600} &
    {\color[HTML]{FF0000} 0.0575}\\
  {\color[HTML]{000000} 6} &
    {\color[HTML]{000000} 54.61} &
    {\color[HTML]{000000} 0.0621} &
    {\color[HTML]{000000} 31.77} &
    {\color[HTML]{000000} 14.38} &
    {\color[HTML]{000000} 2.492} &
    {\color[HTML]{000000} 0.0840} &
    {\color[HTML]{000000} 59.67} &
    {\color[HTML]{000000} 0.0771} &
    {\color[HTML]{000000} 30.49} &
    {\color[HTML]{000000} 11.67} &
    {\color[HTML]{000000} 1.598} &
    {\color[HTML]{000000} 0.0631}
     \\ \hline
  \end{tabular}
  }
  \end{table*}

\begin{figure}[t]
  \centering
  
  \includegraphics[width=3.2in]{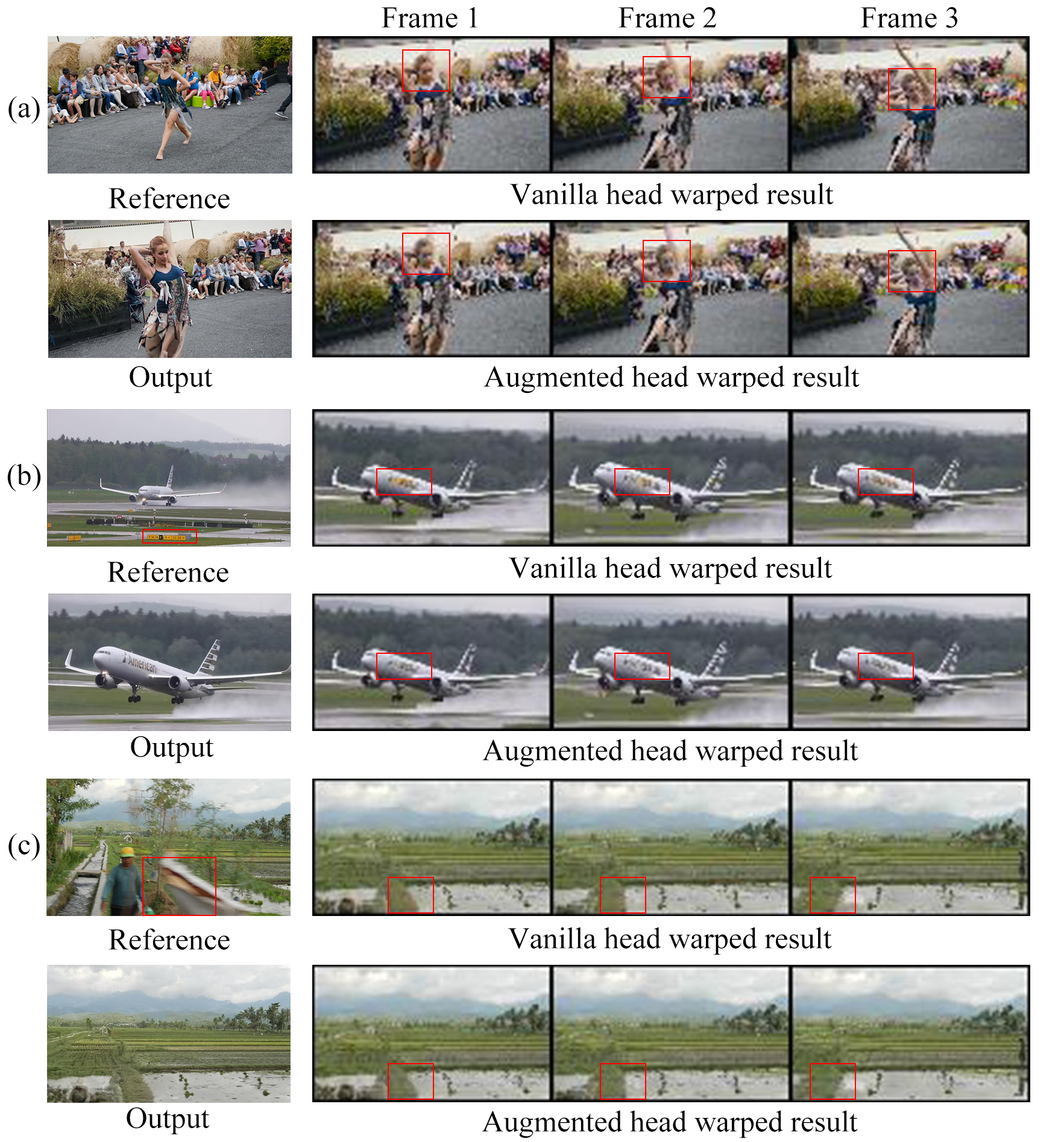}
  
  \caption{Comparison of vanilla and augmented head on warped color result. The vanilla head gives more colorful result in most cases, but as in (b), the word on the plane is correlated to the letters on the yellow sigh. And as in (c), the color of the grass tends to be brown. With more augmented spatiotemporal consistent feature, augmented head gives better color in the last two cases. And as shown in each of the colorized output of frame1, whichever of the two head works better, the Colorization subnet will combine the two and finally generate plausible result}
  \label{doublehead}
\end{figure}

{\bf Visual analysis on double non-local heads.} As described in section \uppercase\expandafter{\romannumeral 3}, the augmented head in double-head non-local operation empirically contains more augmented feature than vanilla head. By visualizing the warped color result of these two head, we can find their functional differences. Firstly, vanilla head performs well under most of the circumstances, and gives more colorful result on foreground object, as illustrated in {\bf Fig. \ref{doublehead}(a)}, the face of the dancer looks plausible in vanilla head's result while the augmented head gives incomplete and faded color. We assume that this is because the vanilla head relies more on texture information, which is less in augmented high-level feature. That makes vanilla head generally works well since most of the correlations can be established by texture characteristics. However, that characteristics make mistakes in some cases. As shown in {\bf Fig. \ref{doublehead}(b)}, the reference image includes a yellow sign under the plane. When the plane gets closer, the word on the side of the plane is dyed yellow, since it have similar texture with the letters in the sign. However, the augmented head doesn't suffer this because of the augmented global information. Besides, we find that vanilla head is sometimes influenced by surrounding. Such as in {\bf Fig. \ref{doublehead}(c)}, the warped result of vanilla head in the red boxes tend to be brown. This is caused by the brown patch between grass and white band in the reference (in the red box). Though some brown in the grass is reasonable in one image, the inconsistent positions of the brown areas across adjacent frames make the results unrealistic. While the augmented head gives plausible spatiotemporal consistent results. Finally, as the output samples in {\bf Fig. \ref{doublehead}}, whichever of the two heads gives better warped color, our Colorization subnet will combine the two and generate plausible results.

{\bf Effect of frame block size.}
Our network can process the video with different frame numbers in a frame block. We evaluate our model with $N$ in range of 1 to 6, as shown in {\bf Table \ref{tab:frames}}. The average processing time for each frame is expressed as DUR (duration). With more frames proceed parallel, the processing time decrease firstly, but increase after $N=3$ in DAVIS dataset and $N=5$ in Videvo dataset. This increase is because of the $O(n^2)$  per layer's complexity in self-attention operation, and the different $N$ for the two datasets to get the lowest DUR may due to their different video frame lengths (approximately 100 frames per video in DAVIS and 300 frames in Videvo). According to {\bf Table \ref{tab:frames}}, as the quality metrics remain relatively stable with $N$, the best performances are achieved at $N=3$ and 5 respectively for DAVIS and Videvo datasets, which are the most efficient.

\begin{figure}[t]
  \centering
  
  \includegraphics[width=3.2in]{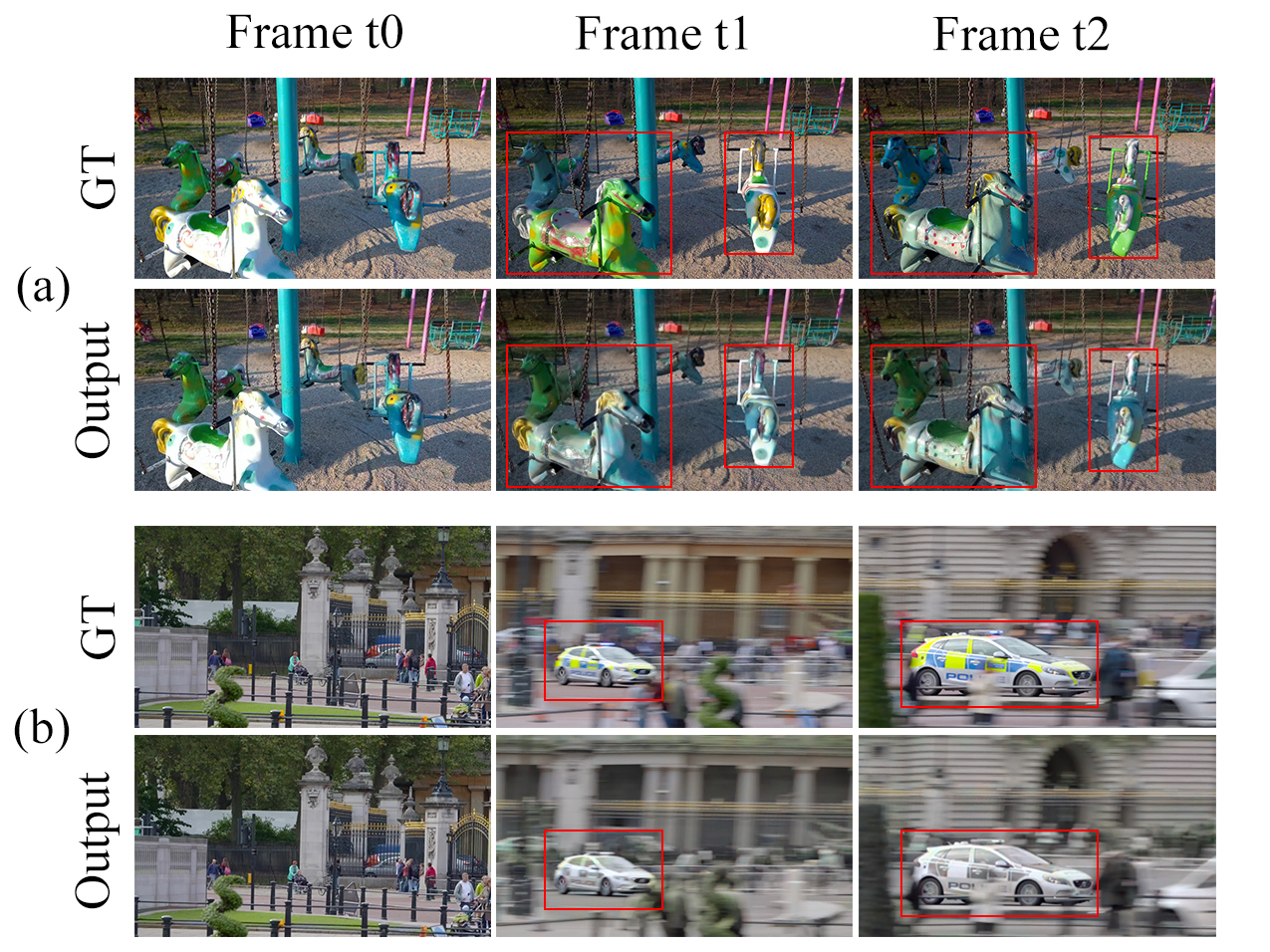}
  
  \caption{Failure cases of our approach. The first case is because of incorrect correspondence to the reference. In {\bf (a)}, as the carousel rotates, frame t1 and frame t2 are very similar to the reference (frame1). In frame t1, the green horse and the white horse moves to the position of the white horse and the blue horse in frame t0. The horses in the same position is extremely similar and are mistakenly correlated, and this lead to inaccurate and inconsistent color. Frame t2 is the same. The second case is due to the presence of the object which is not in reference image. In {\bf (b)}, a police car pulls out of the intersection, and sped along the road. As no appropriate color can be warped from the reference, it is hard to generate colorful results.}
  \label{failure}
\end{figure}

\subsection{Failure cases}
Our method utilizes a reference image as a guidance for colorization. Though this kind of approach generates more colorful and distinct results in most of the cases, it doesn't perform so well under some circumstances, as illustrated in {\bf Fig. \ref{failure}}. The first case is because of incorrect correspondence to the reference. In {\bf Fig. \ref{failure}(a)}, as the carousel rotates, frame t1 and frame t2 become very similar to the reference (frame t0). In frame t1, the green horse and the white horse moves to the position of the white horse and the blue horse in frame t0. The horses in the same position is extremely similar and are mistakenly correlated, and this lead to inaccurate and inconsistent color. Frame t2 is the same situation. The second case is due to the presence of the object which is not in reference image. In {\bf Fig. \ref{failure}(b)}, a police car pulls out of the intersection, and speeds along the road. As no appropriate color can be warped from the reference, it is hard to generate colorful results.

\section{Conclusion}
In this paper, we propose a novel exemplar-based video colorization architecture. We introduce a CNN-Transformer block to model long-term spatial dependency and a double-head non-local operation to play its strength, while the linkage subnet further enhance long-term temporal continuity. Our method produce most colorful and realistic results especially in motion area, but nonetheless has limitation due to strong dependency to reference image. Further, researches will be implemented for more robust and flexible exemplar-based video colorization.


\bibliographystyle{elsarticle-num}
\bibliography{reference}

\end{document}